\pdfoutput=1
\documentclass[11pt]{article}
\usepackage{acl}
\usepackage{times}
\usepackage{latexsym}
\usepackage[T1]{fontenc}
\usepackage[utf8]{inputenc}
\usepackage{microtype}
\usepackage{inconsolata}

\usepackage[british]{babel}
\usepackage[nolist,nohyperlinks]{acronym}
\usepackage{amsmath,amsfonts,amssymb} 
\usepackage{multirow}  % for tables
\usepackage{booktabs}  % for lines in tables
\usepackage{nth}
\usepackage{cleveref}
\usepackage{dirtytalk}  % for quotes
\usepackage{csquotes}

\usepackage{todonotes}

\title{Code-Switched Language Identification is Harder Than You Think}
\author{Laurie Burchell$^1$, Alexandra Birch$^1$, Robert P. Thompson$^2$, Kenneth Heafield$^1$ \\
$^1$Institute for Language, Cognition, and Computation, School of Informatics, \\ University of Edinburgh, 10 Crichton Street, Edinburgh, EH8 9AB, UK \\
$^2$Department of Materials Science \& Metallurgy, Cambridge University,\\27 Charles Babbage Road, Cambridge, CB3 0FS, UK\\
$^1$\texttt{\{laurie.burchell,a.birch,kenneth.heafield\}@ed.ac.uk},
$^2$\texttt{rpt26@cam.ac.uk}}

\begin{acronym}
    \acro{bce}[BCE]{binary cross entropy}
    \acro{ccs}[CCS]{contextual code switching}
    \acro{cs}[CS]{code switching}
    \acro{fpr}[FPR]{false positive rate}
    \acro{lid}[LID]{language identification}
    \acro{llm}[LLM]{large language model}
    \acro{msa}[MSA]{Modern Standard Arabic}
    \acro{nlp}[NLP]{natural language processing}
    \acro{ukri}[UKRI]{UK Research and Innovation}
\end{acronym}

\begin{document}
\maketitle
\begin{abstract}
\Acf{cs} is a very common phenomenon in written and spoken communication but one that is handled poorly by many \acf{nlp} applications. Looking to the application of building \ac{cs} corpora, we explore \ac{cs} \ac{lid} for corpus building. We make the task more realistic by scaling it to more languages and considering models with simpler architectures for faster inference. We also reformulate the task as a sentence-level multi-label tagging problem to make it more tractable. Having defined the task, we investigate three reasonable models for this task and define metrics which better reflect desired performance. We present empirical evidence that no current approach is adequate and finally provide recommendations for future work in this area.
\end{abstract}

\section{Introduction}
\Acf{cs}, or the use of one or more languages within the same utterance \citep{sitaram2019survey}, is a very common phenomenon in written and spoken communication \citep{dogruoz-etal-2021-survey}. However, many \acf{nlp} applications currently struggle to deal with it effectively \citep{calcs-2021-approaches,winata-etal-2023-decades}. An obvious first step in building better systems for \ac{cs} is gathering the data necessary for training effective models, something which is currently lacking for \ac{cs} text \citep{mendels-etal-2018-collecting}. A fundamental part of this process is identifying \ac{cs} in the first place.

In this paper, we look at \ac{cs} \acf{lid} for text and the challenges in getting \ac{cs} \ac{lid} systems to work at scale. Previous shared tasks on \ac{cs} \ac{lid} have produced systems which achieve impressive results \citep{solorio-etal-2014-overview,molina-etal-2016-overview}, albeit limited to two languages. We seek to extend \ac{cs} \ac{lid} systems to work in a realistic setting as part of a corpus building pipeline by scaling up both the number of languages covered and the speed of inference. Our intended use case is mining web text to build \ac{cs} corpora which can then be used as training data in applications aimed at handling \ac{cs}. 

We therefore reformulate \ac{cs} \ac{lid} as a multi-label task where the aim is to assign a set of language labels to each sentence, rather than a word-level or document-level tagging task as in previous work (\Cref{sec:task-def}). We experiment with high-coverage \ac{lid} systems (200+ languages) which are simple enough to scale easily, and investigate three different models as reasonable baseline approaches to the task (\Cref{sec:models}). We test on wide range of \ac{cs} and single-label \ac{lid} test sets aiming to cover as many languages as possible (\Cref{sec:test-sets}), and we choose metrics that better reflect true performance in our multi-label setting than those commonly used for single-label \ac{lid} (\Cref{sec:metrics}). We find that even the best-performing models are still inadequate for identifying \ac{cs} text at scale (\Cref{sec:results}), due to the inherent difficulty of defining \ac{cs} and detecting the intended language(s) in realistic settings. Finally, we make recommendations for future work in this area based on our findings (\Cref{sec:analysis}). To aid future research, we provide code to obtain and transform training and test data, to train all models, and to calculate evaluation metrics.\footnote{\url{https://github.com/laurieburchell/cs-lid-harder-than-you-think}}

\section{Previous work}

\ac{lid} has been an active topic of research for a long time in \ac{nlp} \citep{jauhiainen2019automatic}. Much of the most recent research on this topic has been towards covering more and more languages, with some models claiming to cover over a thousand \citep{brown-2014-non,dunn2020mapping,adebara-etal-2022-afrolid,nllbteam2022language,burchell-etal-2023-open}. However, nearly all general-purpose \ac{lid} systems assume that text is entirely monolingual \citep[e.g.][]{nllbteam2022language} or occasionally that any different languages present occur in discrete chunks \citep[e.g.][]{cld2}. This leads to pipelines where \ac{cs} text is ignored or discarded. 

Previous work on multiple-label \ac{lid} specifically can be split into two main sub-tasks: multilingual \ac{lid}, where the expected input is a document containing discrete monolingual chunks in different languages; and \ac{cs} \ac{lid}, where the expected input is a sentence or short text containing \ac{cs} text. The former task has a longer history and its intended application is to segment web text \citep{baldwin-lui-2010-multilingual,lui-etal-2014-automatic,jauhiainen2015language,kocmi-bojar-2017-lanidenn}. The latter task has received more attention recently including several shared tasks on \ac{cs} \ac{lid}, where the aim was word-level tagging of \ac{cs} text given a known pair of languages \citep{solorio-etal-2014-overview,molina-etal-2016-overview}. However, both tasks have a limited application to web-scale text because they assume that the input is only in a small number of known languages and tend to reply on computationally-expensive, high-capacity models like transformers \citep{vaswani2017attention} or \acp{llm} for classification. We argue that these are not realistic for filtering web crawls since inference is too slow and expensive. 

Finally, we note that despite the wide range of approaches towards monolingual \ac{lid} \citep{jauhiainen2019automatic}, \ac{lid} algorithms are still found to perform poorly in practice compared to test performance, particularly for low resource languages \citep{caswell-etal-2020-language,kreutzer-etal-2022-quality}. This shows that even the simpler task of monolingual high-coverage \ac{lid} remains a challenging problem.

\section{Task definition}
\label{sec:task-def}
We define our task as follows: given a short input text (around sentence length), return a set of codes corresponding to the language(s) it contains. Following \citet{nllbteam2022language}, we output modified ISO 639-3 language codes encoding both the language variety and the script: for example, \texttt{eng\_Latn} 
means English written in Latin text. 

This way of framing the task differs from most previous work on \ac{cs} \ac{lid} by assigning tags on the sentence-level, rather than on the word level as in \citet{solorio-etal-2014-overview,molina-etal-2016-overview}. Less granular labeling like this speeds up inference and so is more practical when using \ac{lid} to build corpora from web-scale text. In addition, we felt that labelling on the sentence-level avoided some of the ambiguity when labelling at the word level. Our model covers many more languages than the previous shared tasks in \ac{cs} \ac{lid} (201 rather than just two) so the search space becomes much larger and less tractable at the word level. In addition, the shared tasks included extra tags aside from the two included languages, covering categories such as named entities, `foreign words', and non-linguistic content like emojis. We wished to avoid this complication since it was not relevant to our aim of dataset building.

\section{Models}
\label{sec:models}
We compare the performance of three models for \ac{cs} \ac{lid}: OpenLID, a pre-existing single-label \ac{lid} model adapted to a multi-label setting \citep{burchell-etal-2023-open}, MultiLID, a novel \ac{lid} model, and Franc, a high-coverage \ac{lid} package.\footnote{\url{https://github.com/wooorm/franc}} The first two models are trained on the same data (OpenLID) to help isolate the effect of the change in architecture. We employ Franc as a comparison point, since it allocates prediction scores in a different way and covers more languages than the other two. In this way, we aim to measure the performance of three reasonable approaches to \ac{cs} \ac{lid}, explore their limitations, and so guide further research.

\subsection{OpenLID}
We adapt the single-label OpenLID \ac{lid} model provided by \citet{burchell-etal-2023-open} to a multi-label setting. We choose this model because it covers a large number of languages with good performance, it scales well to large datasets, and its openly-available training data means we can compare two models trained on the same data and thus eliminate a potential confounding variable.

OpenLID is a \textit{fastText} model \citep{joulin-etal-2017-bag}. The architecture consists of an input sentence vector obtained by averaging word and n-gram embeddings, which is then fed to a simple linear classifier. The output logits are transformed to a probability distribution over the output labels with a softmax activation function. It uses cross entropy loss to update the weights.

We use thresholding to obtain multi-label outputs since this is a standard method to adapt softmax-based classifiers to a multi-label task. This means that rather than returning the label with the maximum probability, we instead return all labels with a predicted probability above some chosen threshold $k$. The classifier may return no labels in the case where no language is predicted a probability above the threshold. It also limits the maximum number of labels to $ \lfloor k^{-1} \rfloor $ because the predicted probabilities for all the classes must sum to one. We set $k=0.3$ so that the classifier can return a maximum of three labels.

Softmax-based classifiers like OpenLID make the implicit assumption that each input should be assigned one and only one label. This is because their output is a probability distribution over mutually-exclusive classes. We therefore experiment with altering the basic architecture of OpenLID to relax this assumption, resulting in the MultiLID model.

\subsection{MultiLID}
We create MultiLID, a novel \ac{lid} model which conceptualises \ac{lid} as a multi-label rather than single-label problem. In this way, we aim to handle both monolingual and \ac{cs} text. There are a range of approaches for multi-label problems \citep{zhang2013review}, but inspired by \citet{stahlberg-kumar-2022-jam}, we explore using \ac{bce} loss: rather than use a softmax activation followed by cross-entropy loss as in OpenLID, MultiLID uses a sigmoid activation plus cross-entropy loss. The effect is that the predicted scores are no longer normalised into a probability distribution so the model can predict multiple classes independently. 

More formally, \ac{bce} is defined as follows. Let $N$ be the number of languages covered by the classifier, $L=[l_1,\dots,l_k,\dots,l_N]^\top$ be the output vector of predicted scores for each language where $l_{k}\in [0,1]$, and $l^*_k \in \{0, 1\}$ be the true label assigned to some input representation $x_k$. The \ac{bce} loss for some particular element $l_k$ is thus:
\begin{multline*}
\textrm{BCELoss}_{l_k} = l^*_k \cdot \log(l_k) + \\
(1-l^*_k) \cdot \log \left( 1-l_k\right)
\end{multline*}
We sum the loss for each element to generate the final loss since we have a sparse output vector.
%This is the sort of thing I'd want a reference for too, but maybe norms differ between fields

When deciding which labels to return, we found that a fixed threshold was ineffective due to the unnormalised scores. Instead, we use the following heuristic to choose the labels to return. We note that the \ac{bce} loss function encourages most scores to be close to zero, and so the mean score is very close to zero. Only some of the scores are significantly above the mean, and these correspond to the labels we want to return. We therefore calculate the mean and standard deviation of the output scores for a particular example, and set a dynamic threshold of two standard deviations above the mean based on empirical results using the LinCE training sets (described in  \cref{sec:test-sets}). We choose the language label with the highest score to ensure we always return a label, and optionally return a second label provided its score exceeds the dynamic threshold. 

We build our model using Python and Pytorch, and we aim to keep it as close to \textit{fastText} as possible by design. We first clean the data and remove emoji and hash symbols, then build the vocabulary from all words seen more than 1000 times, plus the 2- to 5-grams of these words. The input sentence representation vector is formed as a bag of vocabulary embeddings, which is then fed to a linear transformation layer. The output logits are converted to output scores using a sigmoid function. 

We note that our model is trained on single-label rather than \ac{cs} data, even though it is designed to be able to return multiple labels if necessary. We made this decision due to the lack of \ac{cs} training data for most languages, so a practicable \ac{cs} \ac{lid} model would need to be trained without specifically \ac{cs} data for every language pair. Future work could look at exploiting what \ac{cs} data does exist.

\subsection{Franc}
The final \ac{lid} model we use is Franc, a \ac{lid} package covering 414 languages. We include it as an alternative pre-existing model that covers an even larger number of languages than the other two models, and which returns scores that adapt easily to a multi-label setting. Franc is not trained on the same data as the other two models, but rather we use the pre-trained Python model to predict.\footnote{\url{https://github.com/cyb3rk0tik/pyfranc}}

At inference time, Franc returns scores for all languages that use the same script as the input text in decreasing order of probability. These scores are calculated based on the distances of the trigram distributions in the input text and the language model, scaled such that the closest language will have a score of 1. Since we often have short strings in our test sets, we set the minimum valid string length to 1 so Franc always returns a prediction. We choose to return the closest predicted language label plus the second-closest language label provided its predicted score is higher than 0.99 (since this is sufficiently close to still be a valid label). This selection heuristic is based on empirical results on the LinCE training sets (\cref{sec:test-sets}). 

The language labels returned by Franc differ somewhat from those assigned to the test sets. We normalise these using the \texttt{langcodes} Python package,\footnote{\url{https://github.com/rspeer/langcodes}} so if the language code is not among those covered by FLORES-200*, we  find an equivalent tag.\footnote{Specifically, we filter on $\text{\texttt{tag\_distance}}<10$.} If a match exists, we replace the predicted tag with this match; otherwise, we simply return the original prediction. When calculating the metrics, we count all languages not covered by FLORES-200* (described in \cref{sec:test-sets}) as empty tags for ease of computation.

\section{Test sets}
\label{sec:test-sets}
Our aim when choosing test sets was to cover as many \ac{cs} language pairs as possible, despite the limited number of easily-accessible \ac{cs} test sets. We were further hampered by the fact that the OpenLID training data does not include Indian languages written using Roman characters which are some of the most common languages to include in \ac{cs} test sets \citep{aguilar-etal-2020-lince,khanuja-etal-2020-gluecos,winata-etal-2023-decades}. Nonetheless, we source six \ac{cs} test sets which include eight languages, plus a high-coverage monolingual test set. 

We describe all test datasets below and include fuller instructions on how to obtain them in \Cref{sec:data-sourcing}. Most of the datasets we use are annotated with language tags at the token level. To fit with our task, we convert these to sentence-level tags by relabelling the sentence as \ac{cs} if two language labels are present, monolingual if only one is present, and discarding the sentence if it has no language labels (e.g. the sentence only contains named entities or emojis). \Cref{tab:prop_cs} summarises the proportion of \ac{cs} examples in each test set after preprocessing.

\begin{table}[htbp]
    \small
    \centering
    \begin{tabular}{lc}
    Test set & \% \ac{cs} \\
    \midrule
    Turkish--English & 98.9 \\
    Indonesian--English & 93.5 \\
    Basque--Spanish & 59.8  \\
    Spanish--English & 35.2 \\
    Chinese--English & 27.8 \\
    MSA--Egyptian Arabic & 14.5 \\
    FLORES-200* & 0 \\
    \end{tabular}
    \caption{Proportion of \ac{cs} examples in each test set in order of most to least.}
    \label{tab:prop_cs}
\end{table}

\paragraph{Turkish--English dataset} \citet{yirmibesoglu-eryigit-2018-detecting} created a \ac{cs} Turkish--English dataset as part of their work on detecting \ac{cs} for this language pair. The data is sourced from Twitter and the Ek\c{s}i S\"{o}zl\"{u}k online forum, then labelled at the token level as either Turkish or English. After recombining sentences, the dataset consists of 376 lines of data and 98.9\% of the sentences are labelled as \ac{cs}.

\paragraph{Indonesian--English dataset} \citet{barik-etal-2019-normalization} created a \ac{cs} Indonesian--English dataset from Twitter data, where each token in each tweet is annotated with a language tag. After pre-processing, the dataset consists of 825 lines of data and 93.5\% of the sentences are labelled as \ac{cs}.

\paragraph{BaSCo Basque--Spanish corpus} This corpus contains Spanish and Basque sentences sourced from a collection of text samples used in training bilingual chatbots \citep{aguirre-etal-2022-basco}. These sentences were shown to volunteers who were asked to provide a realistic alternative text with the same meaning in Euska\~{n}ol (Basque--Spanish \ac{cs}). The created sentences were checked for validity by a team of annotators. We process this corpus into our test set by extracting all Spanish, Basque, and Euska\~{n}ol utterances present in the final corpus and labelling them using the provided utterance-level language labels. After processing, the dataset consists of 2304 lines of data, of which 59.8\% are labelled as \ac{cs}. 

\paragraph{LinCE Spanish--English and Modern Standard Arabic--Egyptian Arabic} \citet{aguilar-etal-2020-lince} provide a benchmark for linguistic \ac{cs} evaluation, used in previous shared tasks on \ac{cs} \ac{lid} \citep{solorio-etal-2014-overview,molina-etal-2016-overview}. We test on two of its suite of language pairs and tasks, Spanish--English \ac{lid} and \ac{msa}--Egyptian Arabic \ac{lid},\footnote{The other two include transliterated Hindi and transliterated Nepali, neither of which are covered by our \ac{lid} models.} using the validation sets since the test sets are private. These datasets are both sourced from Twitter and are annotated at the word level. After relabelling at the sentence level and filtering, there are 3247 lines of Spanish--English data, of which 35.2\% are marked as \ac{cs}, and 1107 lines of \ac{msa}--Egyptian Arabic data, of which only 14.5\% are marked as \ac{cs}.

\paragraph{ASCEND Mandarin Chinese--English} \citet{lovenia-etal-2022-ascend} created a corpus of conversational Mandarin Chinese--English \ac{cs} speech which is transliterated and labelled by language at the utterance level. We extract the transliterated sentences from the training split of this dataset. After processing, there are 9869 lines of data of which 27.8\% are labelled as containing \ac{cs}.

\paragraph{FLORES-200*} We assess single-label \ac{lid} performance using a subset of FLORES-200, an evaluation benchmark consisting of professional translations from 842 distinct web articles \citep{guzman-etal-2019-flores,goyal-etal-2022-flores}. It includes 3001 sentences for each one of 204 language varieties. Following \citet{burchell-etal-2023-open}, we test on 201 of these taken from the dev-test split, which we refer to as FLORES-200*. We test on this dataset to assess the monolingual performance of our classifier. FLORES-200* consists of 203,412 lines of data after pre-processing.

\section{Measuring performance}
\label{sec:metrics}

The most common metrics for single-label, multi-class problems are precision and recall (defined in \Cref{sec:prec-recall}). However, whilst these metrics give some insight into the functioning of our models, we found them too easy to misinterpret in a multi-label setting. The first reason for this is that precision and recall are undefined when there are no true positive examples of a predicted class in the dataset. This was very common given our high-coverage models, but precision and recall could not detect this key performance issue. Secondly, neither precision nor recall account for true negatives, a key indicator for our application of building web corpora since avoiding spurious labels helps prevent noisy datasets. 

As a consequence of these findings, we decided that precision and recall were not suitable for use as main metrics. Instead, we chose three alternative metrics as a better reflection of the desired downstream performance: exact match ratio, Hamming loss, and \ac{fpr}. These metrics allow direct comparison between our different datasets and are easy to interpret correctly even in a multi-label setup with many classes such as ours. We define and discuss each metric below.

\paragraph{Exact match ratio} This metric is simply that for each sentence $i$ in our dataset of length $N$, we count a correct match if all the predicted labels $(\hat{y})$ match the gold labels $(y)$:
$$
\text{Exact match ratio} = \frac{1}{N} \sum^{N-1}_{i=0} \mathbb{I} (\hat{y}_i = y_i)
$$
The higher the metric, the better. The exact match ratio has the advantage of being easy to understand, but it is a strict measure of success and does not reward partial matches.

\paragraph{Hamming loss} We therefore also report Hamming loss which allows us to both give credit for partial matches and to penalise predicting too many labels. It can be understood as the fraction of wrong labels among the total number of labels, and the smaller the value of the loss the better.  More precisely, let $L$ be the number of classes (languages), $Y_{i,l}$ $(\hat{Y}_{i,l})$ signify the Boolean that the $i^{th}$ example (prediction) is assigned the $l^{th}$ language label, and $\oplus$ denote exclusive-or:
$$
\text{Hamming loss} = \frac{1}{LN} \sum_{i=0}^{N-1}\sum_{l=0}^{L-1} Y_{i,l} \oplus \hat{Y}_{i,l}
$$

\paragraph{False positive rate} Finally, we report the macro-average of \acf{fpr} with respect to each language class, or the ratio of number of examples incorrectly identified as a particular language (false positives, $FP$) to the total number of ground truth negatives (true negatives plus false positives, $TN+FP$). 
$$
\text{False positive rate} = \frac{FP}{TN+FP}
$$
The smaller the \ac{fpr}, the better. Measuring non-relevant predictions is particularly important given our intended application of building web corpora. This is because the internet mostly consists of non-\ac{cs} data, so using a classifier with a high \ac{fpr} on the web will result in a final dataset where most of the content is not relevant \citep{caswell-etal-2020-language}. 

\section{Results}
\label{sec:results}

\begin{table}[htbp]
    \centering
    \small
    \begin{tabular}{clll}
         & MultiLID & OpenLID & Franc \\
         \cmidrule(lr){2-4}
         Exact match $\uparrow$ & 0.861 & \textbf{0.926} & 0.672\\
         Hamming loss $\downarrow$ & 0.00121 & \textbf{0.000694} & 0.00279 \\
         FPR $\downarrow$ & 0.000885 & \textbf{0.000395 }& 0.00123 \\
         Precision $\uparrow$ & 0.879 & \textbf{0.942} &  0.666 \\
         Recall $\uparrow$ & 0.933 & \textbf{0.939} & 0.706 \\
         Mean \# preds. & 1.11 & 1.02 & 1.08 \\
    
    \end{tabular}
    \caption{Results on FLORES-200* test set. We include results using the OpenLID model returning all labels with predicted probability > 0.3 and the top two predictions from Franc with score > 0.99.}
    \label{tab:flores_results}
\end{table}

\begin{table*}[htbp]
    \centering
    \small
    \begin{tabular}{llllllllll}
    & \multicolumn{3}{c}{Exact match $\uparrow$} & \multicolumn{3}{c}{Hamming loss $\downarrow$} & \multicolumn{3}{c}{False positive rate $\downarrow$} \\
    &  MultiLID & OpenLID & Franc & MultiLID & OpenLID & Franc & MultiLID & OpenLID & Franc \\
    \cmidrule(lr){2-4} \cmidrule(lr){5-7} \cmidrule(lr){8-10}
    tur--eng  & \textbf{0.0665} & 0.0213 & 0.00532 & 0.00732 & \textbf{0.00531} & 0.00903 &  0.00206 & \textbf{0.000291} & 0.00119 \\
    ind--eng & \textbf{0.184} & 0.0448 & 0.0182 & \textbf{0.00617} & 0.00680 & 0.00995 &  0.00199 & \textbf{0.00153} & 0.00164 \\
    eus--spa & 0.317 & \textbf{0.360}  & 0.201 & 0.00576 & \textbf{0.00383} & 0.00746 & 0.00213 & \textbf{0.000620}  & 0.00169 \\
    spa--eng   & 0.379 & \textbf{0.417} & 0.146  & 0.00613 & \textbf{0.00451} & 0.00721 & 0.00314 & \textbf{0.00126} & 0.00168 \\
    zho--eng & \textbf{0.508} & 0.507 & 0.301 & 0.00399 & \textbf{0.00386}  & 0.00447 & 0.00197 & 0.00130 & \textbf{0.000332} \\
    arb--arz  & 0.345 & 0.625  & \textbf{0.691} & 0.00631 & 0.00281  & \textbf{0.00242} & 0.00500 & \textbf{0.00174} & 0.00481 \\
    \end{tabular}
    \caption{Main metrics calculated for predictions on the \ac{cs} datasets.}
    \label{tab:all-cs-results}
\end{table*}

\paragraph{FLORES-200* results} We first consider the results on the single-label \ac{lid} test set FLORES-200* in order to provide a point of comparison with later results on \ac{cs} datasets. \Cref{tab:flores_results} shows that the OpenLID classifier achieves the best results for each assessed metric, which is unsurprising given that it is designed as a single-label classifier which covers the languages of FLORES-200*. MultiLID still shows reasonable performance, though Hamming loss and \ac{fpr} are markedly higher. This is likely because MultiLID is more likely to predict multiple labels as shown in the higher number of mean predictions at the bottom of \Cref{tab:flores_results}. The performance for Franc is markedly lower across all metrics, though it should be noted that this model is disadvantaged here by covering far more languages than the other two.

\paragraph{\ac{cs} test sets: main metrics} Moving on to the results for the \ac{cs} test sets, \Cref{tab:all-cs-results} gives the exact match ratio, Hamming loss, and \ac{fpr} for the three assessed models. As shown in \cref{tab:prop_cs}, there is a wide variation between how many sentences labelled as \ac{cs} are present in each test set, from 98.8\% in the Turkish--English dataset to just 14.5\% in the \ac{msa}--Egyptian Arabic dataset. 

In terms of exact label match, MultiLID performs better on the most code-mixed datasets, though the absolute numbers are still much lower compared to single-label performance: compare 0.93 for top-1 OpenLID on FLORES-200* (from \citet{burchell-etal-2023-open})to just 0.06 for MultiLID on the Turkish--English dataset. Similarly, the Hamming loss for all models differs by an order of magnitude compared to OpenLID single-label performance in \Cref{tab:flores_results}, showing that they struggle to label \ac{cs} text correctly.

Franc's algorithm means that it is at a particular disadvantage when dealing with \ac{cs} text, since it bases its prediction partially on the script. In the case of mixed scripts (as in the Chinese--English \ac{cs} data), it often did not return a label at all. This lead to the low \ac{fpr} (better) but low exact match (worse) on this dataset. Additionally, Franc does not cover Arabic dialects including Egyptian Arabic, so it labelled nearly all sentences in the \ac{msa}--Egyptian Arabic dataset as \ac{msa}. This gave it a high exact match score and low Hamming loss compared to the other models since it could not confuse similar Arabic dialects and most of the dataset was actually single-label. However, the fact remains that it does not cover Egyptian Arabic at all, and the higher results here show the limitations of the testing regime.

Notably, the \ac{fpr} of the OpenLID model is lower for every test set compared to the other two models (apart from for Chinese--English as discussed above), sometimes by as much as an order of magnitude. This is despite the fact that exact match and Hamming loss do not differ from MultiLID by that degree. Further investigation shows that this difference comes from the fact that null predictions are often a significant proportion of the OpenLID results, particularly for \ac{cs} sentences. \Cref{tab:empty-preds} gives the percentage of empty predictions by this classifier, which can be as high as 12\% for Spanish--English \ac{cs} sentences. Returning no prediction when no label is assigned a high enough probability does result in a lower \ac{fpr} as the model is not forced to classify the most difficult examples. However, such behaviour may not be desirable when building a corpus since the small number \ac{cs} sentences are more likely to be missed.

\begin{table}[htbp]
    \centering
    \small
    \begin{tabular}{ccc}
     & \% empty & \% c/s empty\\
    \cmidrule(lr){2-3}
    FLORES-200 & 0.092 & - \\
    Turkish--English & 0.798 & 0.806 \\
    Indonesian--English & 9.46 & 9.21 \\
    Basque--Spanish & 0.608 & 0.726  \\
    Spanish--English & 10.6 & 12.0 \\
    Chinese--English & 1.91 & 4.42 \\
    MSA--Egyptian Arabic & 0.632 & 1.24 \\
    \end{tabular}
    \caption{Percentage of empty predictions returned by the OpenLID classifier. The left column gives results over the entire dataset, the right only the \ac{cs} sentences.}
    \label{tab:empty-preds}
\end{table}

\begin{table*}[htbp]
    \centering
    \small
    \begin{tabular}{llllllllll}
    & \multicolumn{3}{c}{Exact match $\uparrow$} & \multicolumn{3}{c}{Hamming loss $\downarrow$} & \multicolumn{3}{c}{False positive rate $\downarrow$} \\
     & MultiLID & OpenLID & Franc & MultiLID & OpenLID & Franc & MultiLID & OpenLID & Franc \\
    \cmidrule(lr){2-4} \cmidrule(lr){5-7} \cmidrule(lr){8-10}
    tur--eng & \textbf{0.0618} & 0.0134 & 0 & 0.00737 & \textbf{0.00535} & 0.00907 & 0.00206 & \textbf{0.000281} & 0.00117 \\
    ind--eng & \textbf{0.153} & 0.00649 & 0.0013 & \textbf{0.00634} & 0.00704 & 0.0103 & 0.00175 & \textbf{0.000968}  & 0.00148 \\
    eus--spa & \textbf{0.0247} & 0.0189 & 0 & 0.00789 & \textbf{0.00563}  & 0.00979 & 0.00226 & \textbf{0.000408} & 0.00171 \\
    spa--eng & \textbf{0.0184} & 0.00613 & 0 & 0.00844 & \textbf{0.00729} & 0.0105 & 0.00259 & \textbf{0.000985} & 0.00190 \\
    zho--eng & \textbf{0.0365} & 0.0164 & 0 & \textbf{0.00618} & 0.00637  & 0.00777 & 0.00107 & 0.000703  &  \textbf{0.000620} \\
    arb--arz & \textbf{0.0994} & 0.0373 & 0 & 0.00766 & 0.00584  & \textbf{0.00535} & 0.00294 & 0.000587  & \textbf{0.000127} \\
    \end{tabular}
    \caption{Main metrics calculated over \ac{cs} sentences only.}
    \label{tab:cs-only-results}
\end{table*}

\paragraph{Performance on \ac{cs} sentences} \Cref{tab:cs-only-results} gives the the main metrics solely on the \ac{cs} sentences in each dataset. MultiLID shows higher performance on exact match for every test sets, but the absolute numbers are still low and there is a notable reduction in performance for the datasets with the least amount of \ac{cs}. This shows that the better numbers in \Cref{tab:all-cs-results} were mostly driven by good results on the single-label sentences. Hamming loss is more mixed but the \ac{fpr} for OpenLID is now an order of magnitude lower across the board. This is due to the larger number of null predictions on \ac{cs} sentences shown in \cref{tab:empty-preds} and discussed above. Similarly, even though Franc has a low \ac{fpr}, it also achieves zero in exact match for nearly every test set, suggesting that the algorithm is not suited to \ac{cs} text. The contrast between the results for exact match and \ac{fpr} demonstrate the need for a suite a metrics which measure different aspects of desired performance.

\paragraph{Precision and recall} We return to the entire \ac{cs} tests sets to calculate precision and recall with respect to each language present. Precision was nearly always very close to one, showing that the predictions that the model did make were very likely to be correct. The only exception to this was Egyptian Arabic, where precision was 0.645 for the OpenLID model, 0.485 for the MultiLID model, and 0 for Franc. This was due to former two models struggling to distinguish between Arabic dialects and a lack of coverage for the latter. 

Recall for each model and language label was much more varied, as can be seen in \Cref{tab:recall}. For the datasets with the highest amount of \ac{cs} (Turkish--English and Indonesian--English), there is a large difference between the recall of the OpenLID model. This suggests that its predictions only contain one of the classes and it is failing to detect the other. The difference is less pronounced for MultiLID, suggesting that it is more likely to detect the presence of the other language. For the other datasets, MultiLID does slightly better in recall overall compared to OpenLID, likely because it returns multiple labels more often. Franc nearly always has lower recall compared to the other two models (apart from the degenerate results for \ac{msa}) though it is important to note that it is disadvantaged by covering more labels.

We draw attention to the (sometimes) relatively high scores for recall and the low scores in \Cref{tab:all-cs-results,tab:cs-only-results}. In particular, we note that considering precision and recall in isolation might lead to the conclusion that using one of these \ac{lid} models in a pipeline would create an adequate \ac{cs} dataset. However, the low exact match scores show just how few of the labels are actually correct, especially for \ac{cs} sentences. This demonstrates the importance here of careful metric selection.

\begin{table}[htbp]
    \centering
    \small
    \begin{tabular}{cccc}
        & \multicolumn{3}{c}{Recall $\uparrow$} \\
        Label & MultiLID  & OpenLID & Franc \\
        \cmidrule(lr){2-4}
        tur  & 0.731 & \textbf{0.952} & 0.435 \\
        eng  &\textbf{0.206}  & 0.032 & 0.027 \\
        \cmidrule(lr){2-4}
        ind  & 0.723  & \textbf{0.727} & 0.227 \\
        eng  & \textbf{0.372 }& 0.066 & 0.063 \\
        \cmidrule(lr){2-4}
        eus  & 0.706  & \textbf{0.858} & 0.459 \\
        spa  & \textbf{0.377}  & 0.312 & 0.128 \\
        \cmidrule(lr){2-4}
        spa  & 0.467  & \textbf{0.469} & 0.193 \\
        eng   & \textbf{0.642}  & 0.560 & 0.211 \\
        \cmidrule(lr){2-4}
        zho  & \textbf{0.792} & 0.695 & 0.467 \\
        eng  & \textbf{0.517} & 0.451 & 0.222 \\
       \cmidrule(lr){2-4}
        arb  & 0.540 & 0.734 & \textbf{0.995} \\
        arz  & \textbf{0.891}  & 0.721 & 0.000 \\
    \end{tabular}
    \caption{Recall with respect to each pair of languages in each \ac{cs} test dataset. Precision is nearly always $\approx 1$.}
    \label{tab:recall}
\end{table}

\paragraph{Number of unique languages predicted} We see from \Cref{tab:num_predicted_langs} that the predictions for all classifiers contain a large number of languages despite there being only two language labels in each test set. This suggests that all three are struggling to form a consistent representation of each language based on the input feature vectors. This may be due to the `confusion' of \ac{cs}, or possibly because of a change of domain from training to test: the training data (at least for OpenLID and MultiLID) is mostly formal text whereas the test data is primarily social media. The predictions for MultiLID contain far more unique languages than those for OpenLID. This is likely because the lack of normalisation in its architecture results in a less strong prior over languages, so it is more likely to predict rarer languages. Franc's predictions nearly always contain far more again, which is probably an artifact of the large number of languages it includes.

\begin{table}[htbp]
    \centering
    \small
    \begin{tabular}{cccc}
     & MultiLID & OpenLID & Franc\\
    \cmidrule(lr){2-4}
    tur--eng & 54 & 11 & 97 \\
    ind--eng & 79 & 27 & 118 \\
    eus--spa & 94 & 50 & 193 \\
    spa--eng & 126 & 86 & 234\\
    zho--eng & 134 & 85 & 225 \\
    arb--arz & 18 & 10 & 8 \\
    \end{tabular}
    \caption{Number of unique languages in the predictions by each model for each \ac{cs} test set.}
    \label{tab:num_predicted_langs}
\end{table}

\section{Analysis}
\label{sec:analysis}
Considering the results as a whole, it is clear that none of the models are adequate for the task of detecting the language(s) of \ac{cs} text. The OpenLID model is not designed to return multiple labels and so misses many examples of \ac{cs} sentences, preferring to label them with a single label or not return a label at all. The MultiLID model has the advantage of being designed to return multiple labels, but the lack of normalisation in the scores means that it is more likely to return spurious labels, as shown in its high \ac{fpr} and larger number of unique languages in the predictions. Franc's algorithm is not suitable for \ac{cs} text since it assumes a single script and is designed for longer pieces of text. In all cases, the low exact match ratios show that if we were building a corpus from this data, we would miss most of the \ac{cs} sentences. 

The performance in general is hampered by one of the inherent problems in \ac{cs} \ac{lid}: the boundaries of \ac{cs} are not defined clearly, even at a linguistic level. In her book on the subject, Gardner states that \ac{cs} \say{is not an entity which exists out there in the objective world, but a construct which linguists have developed to help them describe their data} \citep[p.10]{gardner2009code}. However, both linguists and language users disagree on what should count as \ac{cs}, meaning assigning language labels to text can be an ambiguous task in itself.

We illustrate our point with two contrasting examples. Firstly, this tweet is a fairly straightforward example of a separate English fragment followed by a Spanish fragment: 
\begin{displayquote}
    \texttt{@USER delete that tweet\dots ya lo hize}. 
\end{displayquote}
This makes it easy (for a human annotator) to assign language labels to it. However, there are many more cases of potential \ac{cs} which are much more ambiguous and harder to label. The most common of these is a single-word switch in a sentence \citep[p.30]{gardner2009code}, for example: 
\begin{displayquote}
    \texttt{hoy me siento bien \underline{senior}\dots}.
\end{displayquote}
These short switches complicate labelling for two main reasons. Firstly, there is no clear line between a `borrowed' word, \ac{cs}, and a loan word which is now an accepted part of the language (indeed, loan words start out as \ac{cs}) \citep[p.30]{gardner2009code}. Secondly, short fragments of \ac{cs} can make it difficult to work out which language was intended by the author. This leads to disagreement even amongst expert annotators and consequent `noisy' labels. We also note that the non-standard orthography of social media and informal text can also hamper n-gram based approaches to \ac{lid}.

\subsection{Qualitative analysis: Turkish--English}
As shown in \cref{tab:prop_cs,tab:all-cs-results}, the Turkish--English dataset had the highest proportion of \ac{cs} and the lowest exact match. In light of this, we carried out some qualitative analysis of the OpenLID and MultiLID results to understand what kind of errors the model was making and how these related to the test data.

98.9\% of the test examples are labeled as containing both Turkish and English. Despite this, the most frequent prediction for both models was Turkish alone as shown in \cref{tab:top-five-tur-eng-preds}, which gives the top-five predicted labels by count for each model. There were no cases where both models managed to label a \ac{cs} sentence correctly; in fact, the only time both models gave the gold prediction was for two sentences labeled as Turkish only. We note that for all of the 214 examples where OpenLID predicted Turkish as the sole label, MultiLID gave the same (usually partially correct) prediction. 

\begin{table}[htbp]
    \small
    \centering
    \begin{tabular}{cccc}
        \multicolumn{2}{c}{MultiLID} & \multicolumn{2}{c}{OpenLID} \\
        \cmidrule(lr){1-2} \cmidrule(lr){3-4}
        Predictions & \# & Predictions & \# \\
        \midrule
        Turkish & 341 & Turkish & 214 \\
        English & 6 & English & 25 \\
        English \& Turkish & 5 & English \& Turkish & 23 \\
        C. Tatar \& Turkish & 4 & C. Tatar \& Turkish & 11 \\
        None & 3 & N. Azerb. \& Turkish & 9 \\
    \end{tabular}
    \caption{Top-five languages predicted by the OpenLID and MultiLID models on the Turkish--English test dataset. `C. Tartar' = Crimean Tartar, `N. Azerb.' = North Azerbaijani.}
    \label{tab:top-five-tur-eng-preds}
\end{table}

Based on surface analysis (since none of the authors are Turkish speakers), the examples in the test set appear to be well-formed and there is no clear reason why the models struggled to assign the right labels aside from limitations in the models themselves. We give three representative examples in \cref{tab:tur-eng-examples} where one or both models gave an incorrect prediction (there are no \ac{cs} examples where both models gave a correct prediction). For the first two cases, there is no clear reason why one model predicted two labels and the other only one: both examples consist of mid-sentence switches with relatively long continuous text in both languages. For the final sentence, neither model predicted either of the correct labels. We hypothesise that this is an artifact of the non-standard spelling used in the example, namely repeated letters for emphasis. As \citet{caswell-etal-2020-language} point out, repeated n-grams often cause \ac{lid} systems to fail as an artifact of the models' reliance on character n-gram modelling. Our conclusion from the qualitative analysis is that the \ac{lid} models are not failing to predict correctly in general because of flaws with the test set, but rather because inherent flaws in how the models represent the input.

\begin{table}[htbp]
    \small
    \centering
    \begin{tabular}{p{3.8cm}p{1.3cm}p{1.3cm}}
    & \multicolumn{2}{c}{Predictions} \\
    \cmidrule(lr){2-3}
    Example & OpenLID & MultiLID \\
        \midrule
        bir kahve dükkanında geçen film tadında güzel bir şarkıya ayrılsın gece \textit{falling in love at a coffee shop} & Turkish & English \& Turkish \\
        \midrule
        \textit{haters gon hate players gon play live a life man good luck mic drop} tam bekledigim gibi cikti çok efsane & English \& Turkish & English \\
        \midrule
        deri ceket sezonu acilsinnnnnn \textit{cool kids of} bursaaaaa & Standard Latvian & Latgalian \& Wolof \\
        
    \end{tabular}
    \caption{Examples from the Turkish--English test dataset where the gold labels are `English \& Turkish'. English text is rendered \textit{in italics} to distinguish it from Turkish.}
    \label{tab:tur-eng-examples}
\end{table}

\subsection{Recommendations}
In light of our results and analysis, we have the following suggestions for improving \ac{cs} \ac{lid} over the baseline approaches explored in this paper. 

Firstly, we recommend that researchers consider carefully which metrics they use and in particular how they relate to the downstream performance: for example, the metrics we use in this paper aim to reflect how useful the \ac{lid} model will be for corpus building. We have shown that using metrics common in multi-class tasks for multi-label tasks is easily misleading and that a suite of metrics is necessary to capture performance fully. 

Secondly, any approach should embrace the ambiguity inherent in the task, and aim for a common sense rather than prescriptive definition of what counts as a language \citep[pp.165-7]{gardner2009code}. With respect to NLP, this means considering the task of language labelling in light of the downstream application, rather than assuming that labels are fixed and exclusive. \ac{cs} is too heterogeneous a concept for a `one size fits all' definition to be useful for improving NLP tooling for multilingual users. 

Finally, we believe that the performance of \ac{cs} \ac{lid} depends heavily on the input representation. All of the models we study in the paper rely on n-gram representations, and the poor results across the board suggest that these are not adequate for representing \ac{cs} in actual use. Further work should move beyond n-gram based embeddings so that the input representation could more easily pick up short switches. 

\section{Conclusion}
We explored the task of scaleable \ac{cs} \ac{lid} with the intended use as part of a corpus-building pipeline. We found that three reasonable approaches to the task fell short of the performance required to build useful corpora, demonstrating that the task of realistic \ac{cs} \ac{lid} at scale is far from solved. We recommend that future work choose metrics with care to reflect true performance, understand the ambiguity inherent in \ac{cs}, and fit their definition of \ac{cs} to the intended task rather than enforce a prescriptive definition of the phenomenon.

\section*{Limitations}
The \ac{cs} test sets we use only cover a small fraction of the potential language sets which could be used in multi-lingual communication, and additionally the languages we cover are mostly high-resource (particularly English). Creating more high-quality \ac{cs} datasets for more of the world's languages would be incredibly useful further work.

Though we mitigate some ambiguity by labelling at the sentence- rather than word-level, there is still a level of ambiguity in assigning labels for \ac{lid}. This is particularly apparent for short switches and/or similar languages. Future work could devise better models for ambiguous language labels.

The OpenLID data contains a large amount of skew in the number of training examples per class. This may mean that some classes are more likely to be predicted than others as an artifact of its probability to occur in the training data. Conversely, some languages are more likely to be used for \ac{cs}, particularly English, but our models do not include any explicit prior on which languages are likely to occur in the same utterance. Further research could explore both mitigating unwanted training data biases and including information about which languages are likely to co-occur. 

\section*{Ethics Statement}
Using social media data to build corpora needs to be done with care so as not to violate users' rights to privacy. The \ac{cs} test sets based on social media in this work have been anonymised and we provide links to the data for further research rather than hosting the files ourselves; this is to help control distribution of the data. We hope that by creating more \ac{cs} datasets, \ac{nlp} technologies become accessible for more people in their preferred language and register of communication.

Updates to the FLORES-200 dataset have raised issues both with the reliability of the test sets and the choice of language labels.\footnote{\url{https://oldi.org/}} We have used the labels used by \citet{burchell-etal-2023-open} in this paper to allow comparison with previous work, but future work should incorporate any updates to the FLORES+ test set. This not only increases the reliability of the test sets, but also incorporates more of the exonyms preferred by the users of the languages themselves.

\section*{Acknowledgements}
This work was supported in part by the \ac{ukri} Centre for Doctoral Training in Natural Language Processing, funded by the \ac{ukri} (grant EP/S022481/1) and the University of Edinburgh, School of Informatics and School of Philosophy, Psychology \& Language Sciences. This work was also funded by \ac{ukri} under the UK government’s Horizon Europe funding guarantee (grant numbers 10052546 and 10039436).

The experiments in this paper were performed using resources provided by the Cambridge Service for Data Driven Discovery (CSD3) operated by the University of Cambridge Research Computing Service (www.csd3.cam.ac.uk), provided by Dell EMC and Intel using Tier-2 funding from the Engineering and Physical Sciences Research Council (capital grant EP/P020259/1), and DiRAC funding from the Science and Technology Facilities Council (www.dirac.ac.uk). 

Special thanks to Henry Coxe-Conklin for his help with the earliest iterations of this paper, to Patrick Chen for his help with Chinese languages, to Nikita Moghe for coming up with the title, and to Emelie Van De Vreken for proof-reading. We would also like to thank the anonymous reviewers who gave us detailed and helpful feedback.

\bibliography{anthology,custom}
\newpage

\appendix

\section{Data sourcing}
\label{sec:data-sourcing}
We provide instructions on how we obtained all datasets used in this paper to aid future work. These are correct at the time of writing; we cannot guarantee that datasets will be available in the future.
\begin{itemize}
    \item OpenLID training dataset: downloaded from \url{https://github.com/laurieburchell/open-lid-dataset}.
    \item FLORES-200 benchmark: downloaded from \url{https://github.com/facebookresearch/flores/blob/main/flores200}.
    \item Turkish--English dataset: fill out and email requisition form at \url{http://tools.nlp.itu.edu.tr/Datasets}.
    \item Indonesian--English dataset: emailing lead author \citep[see][for contact details]{barik-etal-2019-normalization}.
    \item BaSCo Basque--Spanish dataset: \texttt{valid\_utterances.json} downloaded from \url{https://github.com/Vicomtech/BaSCo-Corpus}.
    \item LinCE \ac{lid} benchmark: validation data sourced from \url{https://huggingface.co/datasets/lince}.
    \item ASCEND Chinese--English dataset: training data sourced from \url{https://huggingface.co/datasets/CAiRE/ASCEND}.
\end{itemize}

\section{Precision and recall}
\label{sec:prec-recall}
Let $TP$ be the count of true positives, $FP$ be the count of false positives, and $FN$ be the count of false negatives. Then 
\begin{align*}
    \text{precision}&=\frac{TP}{TP+FP}\:, \\ 
    \text{recall}&=\frac{TP}{TP+FN}
\end{align*}

\end{document}